\documentclass[final]{cvpr}
\usepackage{times}
\usepackage{graphicx}
\usepackage{amsmath}
\usepackage{amssymb}
\usepackage[pagebackref,breaklinks,colorlinks]{hyperref}
\usepackage{subfig}

\usepackage{multirow}

\usepackage{algorithm}
\usepackage{algpseudocode}
\usepackage[accsupp]{axessibility} 

\def\cvprPaperID{5085} 
\def\confName{ICCV}
\def\confYear{2023}
\graphicspath{{./fig/}}

\newcommand{\minisection}[1]{\vspace{-0.05in} \paragraph{#1}}

\begin{document}

\title{Masked Autoencoders are Efficient Class Incremental Learners}

\author{
Jiang-Tian Zhai $^{1}$  \quad
Xialei Liu $^{1, }$\thanks{Corresponding author (xialei@nankai.edu.cn)} \quad
Andrew D. Bagdanov $^2$ \quad
Ke Li $^{3}$ \quad
Ming-Ming Cheng $^1$ \\
$^1$ VCIP, CS, Nankai University \quad
$^2$ MICC, University of Florence \quad
$^3$ Tencent Youtu Lab
}

\maketitle

\begin{abstract}
  Class Incremental Learning (CIL) aims to sequentially learn new classes 
  while avoiding catastrophic forgetting of previous knowledge. 
  We propose to use Masked Autoencoders (MAEs) as efficient learners for CIL. 
  MAEs were originally designed to learn useful representations 
  through reconstructive unsupervised learning, 
  and they can be easily integrated with a supervised loss for classification. 
  Moreover, MAEs can reliably reconstruct original input images from 
  randomly selected patches, 
  which we use to store exemplars from past tasks more efficiently for CIL. 
  We also propose a bilateral MAE framework to learn from image-level 
  and embedding-level fusion, 
  which produces better-quality reconstructed images and 
  more stable representations. 
  Our experiments confirm that our approach performs better than 
	the state-of-the-art on CIFAR-100, ImageNet-Subset, and ImageNet-Full. 
  The code is available at \url{https://github.com/scok30/MAE-CIL}.
\end{abstract}

\section{Introduction}
\label{sec:intro}
\thispagestyle{empty}
Deep learning has had a broad and deep impact on most computer vision tasks 
over the last ten years. 
Given the way humans learn continually in their lifespan, 
it is natural to expect models also to be able to accumulate knowledge 
and build on past experiences to adapt to new tasks \textit{incrementally}. 
The real world is very dynamic, 
leading to varying data distributions over time, 
while deep models tend to catastrophically forget old tasks 
when learning new ones~\cite{mccloskey1989catastrophic}.

Class Incremental Learning (CIL) aims to learn 
new classification tasks sequentially while avoiding catastrophic forgetting
\cite{belouadah2021comprehensive,masana2020class}. 
CIL approaches can be roughly divided into 
three categories~\cite{delange2021continual}, 
\ie, {\emph{Rehearsal-based methods}~\cite{isele2018selective,rebuffi2017icarl,rolnick2019experience}, 
\emph{Regularization-based methods}~\cite{kirkpatrick2017overcoming,li2016learning}, 
and  \emph{Architecture-based methods}~\cite{aljundi2017expert,mallya2018piggyback,mallya2018packnet}.} 
Among them, \emph{Rehearsal-based methods} achieve state-of-the-art 
performance by storing exemplars from past tasks or generating 
synthetic samples for replay.

Normally, only a fixed size of memory is allowed during incremental learning.
Therefore, it limits the stored exemplars from past tasks. 
Other works exploit generative networks
\cite{wu2018memory,xiang2019incremental,zhai2019lifelong} (e.g., GANs) 
to synthesize samples from old tasks for replay. 
Although they can generate replay data to mitigate forgetting, 
a typical drawback is the quality of generated images, 
and that forgetting can also happen in generative models. 
In this work, we introduce Masked Autoencoders (MAEs)~\cite{he2022masked} 
as a base model to replay. 
It allows efficient exemplar storage by only requiring a small subset 
of patches to reconstruct whole images. 
Therefore, we can store more exemplars with the same amount of limited memory 
as other exemplar-based approaches. 
Compared to previous generative methods, replay by MAE is more stable because 
it uses partial cues to infer global information,
which is task-agnostic and suffers less forgetting across tasks. 
This relieves the unstable generation effect of GANs across tasks 
with stationary image patches. 



\begin{figure}[t]
	\small
	\centering
	\includegraphics[width=\linewidth]{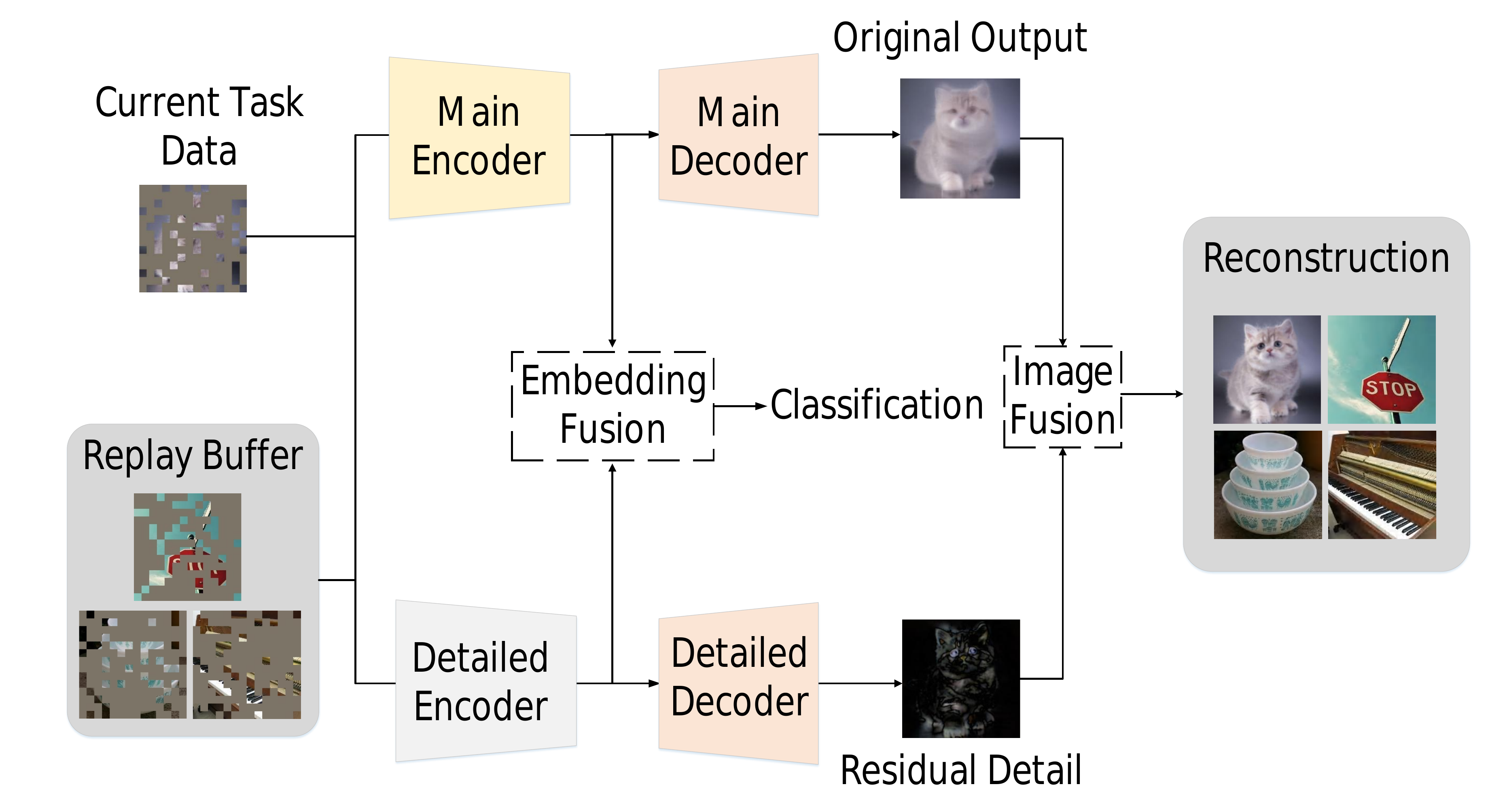}
	\caption{Our proposed bilateral MAE for efficient CIL. 
		The Replay Buffer contains random patches selected from past task images, 
		which is more efficient than storing whole images. 
		Combining these with masked input data from the current task, 
		the MAE simultaneously learns to classify and reconstruct images 
		from the masked input. 
		To further improve the reconstructed image quality and learned representations, 
		embedding-level and image-level fusion is used to learn more stable 
		representations and more detailed reconstructions for CIL. 
	}\label{Fig:fig1}
\end{figure}


%

Masked Autoencoders (MAEs)~\cite{he2022masked} were initially proposed to 
learn better feature representations in self-supervised learning scenarios. 
In this work, we see it as efficient incremental learners and propose a novel 
bilateral transformer architecture for efficient exemplar replay in CIL. 
Our main idea is simple: by randomly masking patches of input images 
and training models to reconstruct the masked pixels, 
MAEs can provide a new form of self-supervised representation learning 
for CIL and thus learn more generalizable representations essential for CIL. 
In addition, leveraging a supervised objective with classification 
labels benefits unsupervised MAE in training efficiency and 
model robustness~\cite{liang2022supmae}. 
Masked inputs can also serve as strong classification regularization by 
only providing a random subset of data.

When learning new tasks, the MAE can coarsely reconstruct images 
from sparsely sampled patches from exemplars. 
This process enables the framework to generate reconstructed replay data, 
but two problems remain: 
(i) the generated images tend to have less detailed and 
less realistic textures, which reduces data diversity for replay; and 
(ii) on the embedding-level, the linear classifier lacks information 
from low-level features. 
Therefore, we introduce a bilateral MAE framework with image-level 
and embedding-level fusion for CIL 
(see \figref{Fig:fig1} for a schematic overview). 
Fusing a complementary detailed and reconstructed image 
alleviates catastrophic forgetting by enriching the insufficient replay data 
with detailed, high-quality data distributions. 
Embedding-level fusion from the two branches also maintains stable and 
diverse embeddings, and our framework can thus achieve a better 
trade-off between plasticity and stability.

The main contributions of our bilateral MAE framework are threefold:
\begin{itemize}
  \item We introduce an MAE framework for efficient incremental learning 
	that incorporates benefits from both self-supervised reconstruction 
	and data generation for replay.
  \item To further boost the quality of reconstructed images and learning efficiency, 
	we design a novel bilateral MAE with two complementary branches 
	for better-reconstructed images and regularized representations.
  \item Our approach achieves state-of-the-art performance under 
	different CIL settings on CIFAR-100, ImageNet-Subset, and ImageNet-Full.
\end{itemize}

\section{Related Work}
\label{sec:rw}


\minisection{Incremental learning.}
Various methods have been proposed for incremental learning 
in the past few years~\cite{delange2021continual,belouadah2021comprehensive}. 
Recent works can be coarsely grouped into three categories: 
replay-based, regularization-based, and parameter-isolation methods. 
Replay-based methods mitigate the task-recency bias by replaying training 
samples from previous tasks. 
In addition to replaying samples, BiC~\cite{wu2019large}, 
PODNet~\cite{douillard2020podnet}, and iCaRL~\cite{rebuffi2017icarl} 
apply a distillation loss to prevent forgetting and enhance model stability. GEM~\cite{lopez2017gradient}, AGEM~\cite{AGEM}, and MER~\cite{MER} exploit past-task exemplars by modifying gradients on current training samples to match old samples. Rehearsal-based methods may cause models to overfit stored samples.

Pseudo-replay methods reconstruct the old data for replay. 
MeRGANs~\cite{wu2018memory} use conditional GANs to balance the generation 
of old and current samples. 
Besides, dreaming-relevant methods like DeepInversion~\cite{yin2020dreaming} 
and AlwaysBeDreaming~\cite{smith2021always} 
exploit backward signals to generate images similar to the original datasets.

Regularization-based approaches, such as LwF~\cite{li2016learning}, 
EWC~\cite{kirkpatrick2017overcoming}, and DMC~\cite{zhang2020class}, 
offer methods to learn better representations 
while leaving enough plasticity for adaptation to new tasks. 
Parameter-isolation methods~\cite{mallya2018packnet,xu2018reinforced} 
use models with different computational graphs for each task. 
With the help of growing models, new model branches mitigate catastrophic 
forgetting at the cost of more parameters and computational costs.  

\begin{figure*}[t]
	\small
	\centering
	\includegraphics[width=\linewidth,height=0.28\textheight]{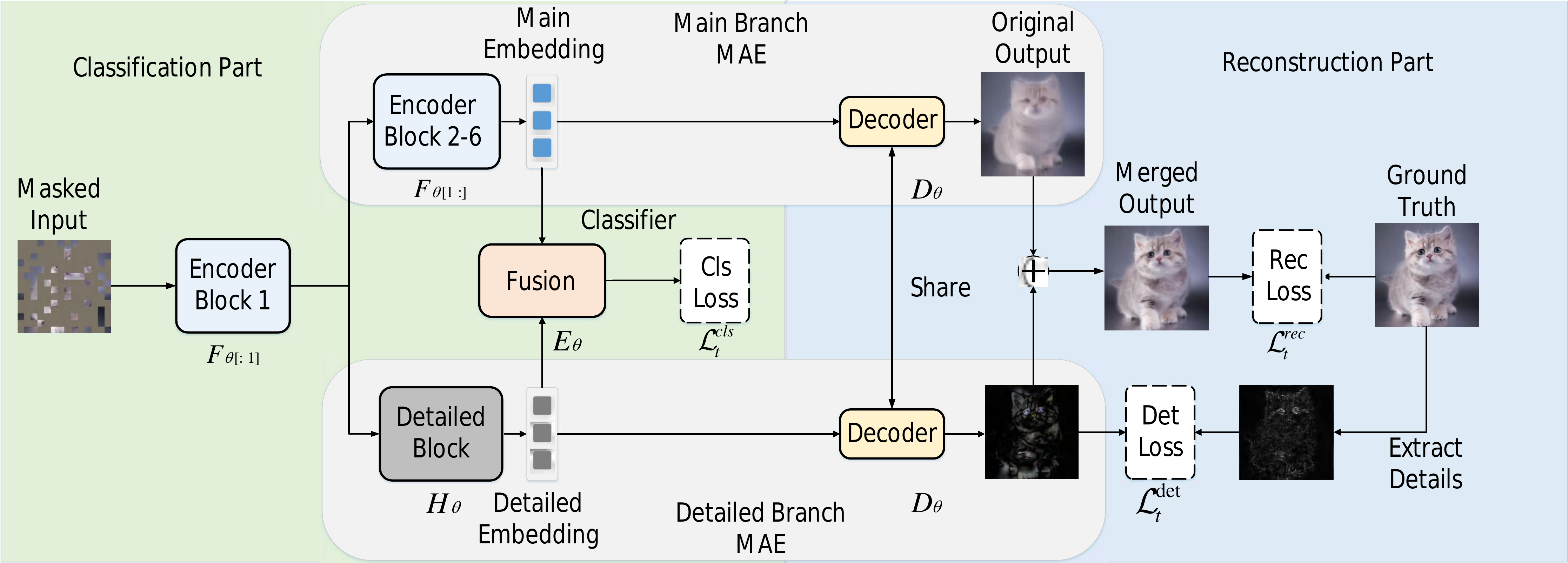}
	\caption{Overall framework of our bilateral MAE for CIL. 
	  The masked input is passed through two branches with 
		embedding-level fusion for classification and 
		image-level fusion for reconstruction. 
		Full images can be generated from a small sample of input patches 
		and the reconstructed images can be used as replay. 
  }\label{Fig:main}
\end{figure*}


\minisection{Self-Supervised Learning.}
Self-supervised learning
\cite{doersch2015unsupervised,wu2018unsupervised,noroozi2016unsupervised} 
has been shown to help models learn generalizable features, 
which makes it natural to consider its application to incremental learning. 
Early works used  pretext tasks like 
patch permutation~\cite{noroozi2016unsupervised} or 
rotation prediction~\cite{gidaris2018unsupervised}.
Contrastive learning approaches model the pairwise similarity and 
dissimilarity between samples~\cite{chen2020simple}. 
By comparison, MAEs~\cite{he2022masked} learn feature representations 
by reconstructing images from a masked version of inputs. 

There are a few works proposing self-supervision for class incremental learning.
PASS~\cite{zhu2021prototype} incorporates 
rotation prediction~\cite{gidaris2018unsupervised} 
to learn representations transferable across tasks. 
DualNet~\cite{pham2021dualnet} uses Barlow Twins~\cite{zbontar2021barlow} 
to introduce a ``slow'' task that regularizes the ``fast'' 
incremental learning. 
In this paper, we explore the framework generates data for replay 
while applying semantic and detailed-level self-supervision, 
which mitigates forgetting via richer replay data 
and more generalizable features.

\section{Continual MAE for CIL}
\label{sec:me}

In this section, 
we first define the class incremental learning problem and the basic MAE model. 
Then we introduce our incremental learning framework and 
the bilateral MAE architecture it is based on.  

\subsection{Preliminaries}

\minisection{The class incremental learning problem.}
CIL aims to sequentially learn tasks with new classes while preventing 
or alleviating forgetting of old tasks. 
At a specific phase of learning task $t \in \{1,2,...,T\}$, 
model training can exploit only data from the current task 
$\{(x_i^t, y_i^t)\}$, 
where $x_i^t$ denotes image $i$ in task $t$ and 
$y_i^t$ its corresponding class label. 
A CIL model typically consists of a feature extractor $F_\theta$ 
and a common classifier $G_\phi$ which grows at each new task. 
At task $t+1$, $C_{t+1}$ new classes are added to $G_\phi$. 
The feature extractor $F_\theta$ first maps the input $x$ to 
a deep feature vector $z=F_\theta(x)\in \mathbb{R}^d$, 
where $d$ is the dimension of the output feature representation, 
and then the unified classifier $G_\phi(z) \in \mathbb{R}^{C_{1:t}}$ 
produces a probability distribution over classes $C_{1:t}$ 
which is used to make predictions on input $x$. 

When training task $t$, the model aims to minimize losses on the current task 
without degrading performance on previous tasks. 
A common technique for mitigating forgetting is to retain a small buffer 
of training samples from previous tasks. 
Let $\epsilon$ be this buffer of previous task samples. 
One essential problem for CIL is the limited amount of replay data. 
Compared to the full data of current task $t$, 
only a few samples of old task classes are available 
(20 samples per class is a common setting), 
which causes imbalanced training between new and old tasks.


\minisection{An MAE framework for classification.}
%
An MAE first crops input image $x$ into non-overlapping patches, 
and we denote the number of patches in the full image $x$ as $N_f$. 
After patchification, the MAE randomly masks a proportion $r \in [0, 1]$ 
of the $N_f$ patches, 
leaving only $N = \lfloor N_f \times (1-r) \rfloor$ patches. 
Then these sampled $K \times K$ pixel patches are mapped to 
a visual embedding of dimension $D$ using an MLP. 
After concatenating with a class token, the result is a tensor of size 
$\mathbb{R}^{(N+1) \times D}$. 
After positionally encoding the original patch locations, 
this input is passed to the MAE transformer encoder. 
This operation maintains the same shape of embedding. 
The output class token embedding can be used for classification 
with a cross-entropy loss $\mathcal{L}_{t}^{\mathrm{cls}}$, 
as shown in \figref{Fig:main}.

For the MAE Decoder, learnable mask tokens are inserted into the embeddings 
in place of the masked patches, 
and the shape of output from the MAE Encoder changes from 
$\mathbb{R}^{(N+1)\times D}$ to $\mathbb{R}^{(N_f+1)\times D}$. 
Although the decoder is not used for classification, 
it helps the network back-propagate image-level reconstruction supervision 
to the embedding-level. 
This stabilizes the image embedding and helps the optimization process. 
Also, the reconstructed images after decoding provide richer, 
higher-quality replay data. 
To limit computation, we use a single-layer transformer block for the decoder. 
The additional classification loss after the encoder speeds up convergence 
and improves reconstruction efficiency during training. 
The mean squared error between the input image $x$ and reconstructed image 
$\hat{x}$ is used as the reconstruction loss function 
$\mathcal{L}_{t}^{\text{rec}}(x, \hat{x})$.



\subsection{Efficient Exemplar Storage with MAEs}
\label{subsec:exemplar}

After training each task, we save small sample images and apply random masking. 
Keeping the same storage size can save more replay data per class 
since each sample occupies less space. 
For example, a masking ratio of 0.75 allows us to save 4$\times$ 
the number of (reconstructible) samples compared to 
conventional replay methods. 

Let $S$ and $P$ denote the size of the image and patch, respectively. 
Our encoder patchifies the input image into 
$\frac{S}{P}\times \frac{S}{P}$ patches. 
We save the 2D index $(i,j)$ for each patch that is not masked. 
We need only one byte to save these indexes since their range is less than 255. 
The two additional bytes for saving this 2D index are negligible compared 
to the saved image patches. 
A masking ratio 0.75 on a 224 $\times$ 224 image requires only 36.75KB 
of storage. 
For $P=16$, the number of saved patches is 
$(1-0.75)\times(\frac{224}{16})^2=49$, and the storage of indexes is only 98B.



\subsection{Bilateral MAE Fusion}
\label{subsec:fusion}

To further boost reconstruction quality and embedding diversity, 
we propose a two-branches MAE to learn global and detailed 
classification and image reconstruction knowledge. 
We illustrate the overall framework in \figref{Fig:main}. 
Bilateral fusion at the embedding-level aims to improve 
representations diversity. 
Reconstruction learning at the image-level yields high-quality replay data 
and stable self-supervision for CIL. 



\newcommand{\addex}[1]{\includegraphics[width=0.24\linewidth]{example/#1.jpg}}
\begin{figure}[t]
  \centering
  \setlength\tabcolsep{1pt}
  \renewcommand\arraystretch{0.6}
	\footnotesize
  \begin{tabular}{cccc}
    Mask ratio=40\%&Mask ratio=75\%&Difference&High-frequency\\
    \addex{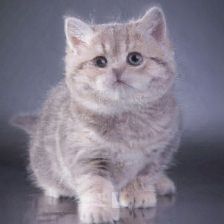}&\addex{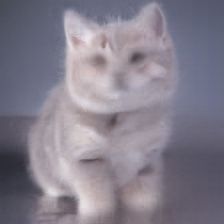}&\addex{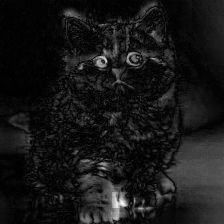}&\addex{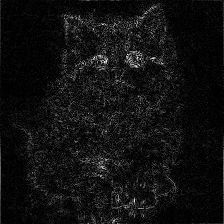}\\
    $\hat{x}_1$ (Eq. \ref{eqn:x1}) & $\hat{x}_2$ (Eq. \ref{eqn:det}) & $\hat{x}_1 - \hat{x}_2$ (Eq. \ref{eqn:l-det}) & $x''$ (Eq.~\ref{eqn:high-freq}) 
  \end{tabular}
	\caption{An example of extracting detailed images with reconstructed results 
	  from different masking ratios $r_1$ and $r_2$. 
		The third image comes from the difference between the previous two, 
		and the last image is the extracted high-frequency component from 
		the third image.
	}\label{fig:ex}
\end{figure}

\minisection{Embedding fusion.} 
In the following we use $F_{\theta[:1]}$ and $F_{\theta[1:]}$ to denote 
our transformer encoder's first and following blocks.  
Let $H_\theta$ and $E_\theta$ represent the detailed block and embedding 
fusion module in \figref{Fig:main}, 
which are standard MLP layers and attention blocks, respectively. 
The classification loss is computed as:
\begin{eqnarray}
  f &=& F_{\theta[:1]}(\mathrm{mask}(x,r))\\ 
  z &=& E_{\theta}(F_{\theta[1:]}(f),H_{\theta}(f))\\
  \mathcal{L}_{t}^{\mathrm{cls}}(x,y) &=& \mathcal{L}_{\mathrm{ce}}(G_\phi(z),y),
\end{eqnarray}
where $\mathrm{mask}(x,r)$ denotes applying random masking with ratio $r$ on image $x$,
$f$ is the embedding extracted by the first encoder block, which is the input to the two branches of our Bilateral MAE, and $G_\phi(z)$ is the estimated class distribution used for in the cross entropy loss.

\begin{algorithm}[!t]
	\renewcommand{\algorithmicrequire}{\textbf{Input:}}
	\renewcommand{\algorithmicensure}{\textbf{Output:}}
	\caption{Pseudocode of our Bilateral MAE.} 
	\label{alg:mae} 
	\begin{algorithmic}[1] 
		\Require 
		The number of task $T$, training samples $D_t = \{(x_i, y_i)\}_t$ of task $t$, model $\Theta^0$, replay buffer $\epsilon$, and masking ratios $r$, $r_1$, $r_2$.
		\Ensure 
		model $\Theta^T$
		\For {$t\in$ $\{1,2,...,T\}$}
		\State $\Theta^t$ ← $\Theta^{t-1}$
		\State $R_t$ ← ReconstructOldSamples ($\epsilon_t$, $r$)
		\While {not converged}
		\State $(x, y)$ ← Sample ($R_t, D_t $)
		\State ($\mathcal{L}_{t}^{\mathrm{cls}}$, $\mathcal{L}_{t}^{\mathrm{rec}}$) ← BilateralMAE ($x, y$)
		\State ($\hat{x}_1, \hat{x}_2$) ← MaskAndReconstruct ($x$, $r_1$, $r_2$)
		\State $\mathcal{L}_{t}^{\mathrm{det}}$ ← ComputeDetailLoss $(\hat{x}_1, \hat{x}_2)$\
		\State train $\Theta^t$ by minimizing $\mathcal{L}_t$ from Eq.~\ref{eqn:all_loss}
		\EndWhile
		\EndFor
	\end{algorithmic} 
\end{algorithm}

\begin{figure*}[!t]
  \centering
  \includegraphics[width=.8\linewidth]{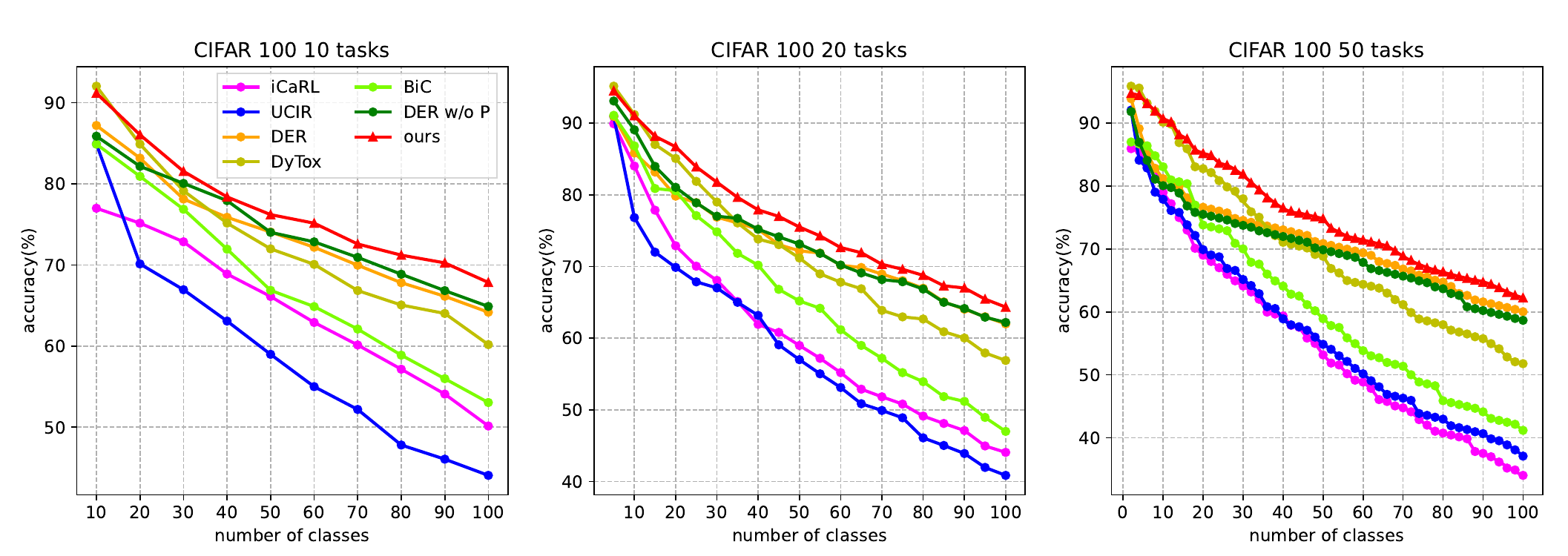}
  \vspace{-5pt}
	\caption{ Performance evolution over incremental tasks on CIFAR-100 on 
	  10-, 20- and 50-task scenarios. 
	}\label{fig:cifar}
\end{figure*}

\begin{table*}[t]
	\centering
	\small
	\renewcommand{\arraystretch}{1.1}
  \setlength\tabcolsep{1.2mm}

	\begin{tabular}{ccccccccccccccc} 
\hline
	  \multirow{2}{*}{Method} & &\multicolumn{3}{c}{$N$=10}&&\multicolumn{3}{c}{$N$=20}&&\multicolumn{3}{c}{$N$=50} \\
	  \cline{3-5} \cline{7-9} \cline{11-13}
	   
	   &&Avg $\uparrow$&Last $\uparrow$&$F\downarrow$&&Avg $\uparrow$&Last $\uparrow$&$F\downarrow$&&Avg $\uparrow$&Last $\uparrow$&$F\downarrow$\\
	   \hline
iCaRL~\cite{rebuffi2017icarl}&&65.27&50.74&31.23&& 61.20&43.75&32.40&&56.08&36.62&36.59
\\
UCIR~\cite{hou2019learning}&&58.66&43.39&35.67&& 58.17& 40.63&37.75&&56.86&37.09&38.13\\
BiC~\cite{wu2019large}&&68.80&53.54&28.44&&66.48&47.02&29.30&&62.09&41.04&34.27\\
PODNet~\cite{douillard2020podnet} &&58.03&41.05&41.47&&53.97&35.02&36.70&&51.19&32.99&40.42\\
DER w/o P~\cite{yan2021dynamically}&&75.36&65.22&15.02&&74.09&62.48&23.55&&72.41&59.08&26.73\\
DER~\cite{yan2021dynamically} &&74.64&64.35&15.78&&73.98&62.55&23.47&&72.05&59.76&26.59\\
DyTox~\cite{douillard2022dytox}&&  75.47&62.10&15.43&& 75.10& 59.41 &21.60&& 73.89& 57.21& 24.22\\ \hline
Ours&&  \textbf{79.12} &\textbf{68.40}& \textbf{12.17}&& \textbf{78.76}& \textbf{65.22}& \textbf{14.39}&& \textbf{76.95}& \textbf{63.12}& \textbf{18.34}\\
\hline
	\end{tabular} 
	\caption{Results on CIFAR-100 in average accuracy (\%), last phase accuracy (\%), and forgetting F (\%) on 10-, 20- and 50-task scenarios.}
	\label{tab:cf}
  \end{table*}

\minisection{Image fusion with detailed loss.}
For the detailed head and corresponding loss, 
we discovered that working in the frequency domain makes it easier 
for the network to attend to high-frequency details, 
which is exactly what the detailed branch should reconstruct. 
We define a frequency-masking function $M(\cdot)$ that converts 
its argument (an image patch) to the frequency domain, 
then masks out low frequencies using a circular mask around the origin.
As shown in \figref{Fig:main}, the MAE decoder is shared by the two branches 
of our model since they have similar reconstruction tasks, 
as well as input and output shapes.
Let $D_\theta$ be this shared decoder, 
then the image-level outputs of the two branches and the reconstruction loss
$\mathcal{L}_{t}^{\mathrm{rec}}$ are:
\begin{eqnarray}
f &=& F_{\theta[:1]}(\mathrm{mask}(x,r))\\
x^{\prime} &=& D_{\theta}(F_{\theta[1:]}(f))\\
\label{eqn:recl}
x^{\prime\prime} &=& \mathrm{ifft2} (M(D_{\theta} (H_{\theta}(f))))
\label{eqn:high-freq}
\\
\hat{x} &=& x^{\prime} + x^{\prime\prime} \\
\mathcal{L}_{t}^{\mathrm{rec}} &=& \mathcal{L}_{\mathrm{mse}}(x, \hat{x}),
\end{eqnarray}
where $x^{\prime}$ and $x^{\prime\prime}$ are the main and residual 
detailed outputs, respectively, 
and $\mathrm{ifft2}$ is the inverse Fast Fourier Transform.


The detailed loss $\mathcal{L}_{t}^{\mathrm{det}}$ also makes use of the 
frequency-masking function $M$ to compare the output of the detailed branch 
and the difference between two MAE reconstructions of the input image:
\begin{eqnarray}
\label{eqn:x1}
  \hat{x}_1 &=& D_{\theta}(F_{\theta}(\mathrm{mask}(x,r_1)))\\
    \label{eqn:det}
  \hat{x}_2 &=& D_{\theta}(F_{\theta}(\mathrm{mask}(x,r_2)))\\
  \label{eqn:l-det}
  \mathcal{L}_{t}^{\mathrm{det}} &=& ||M(D_{\theta} (H_{\theta}(f))) - M(\hat{x}_2-\hat{x}_1)||_1,
\end{eqnarray}
where $\hat{x}_1$ and $\hat{x}_2$ are two reconstructed images 
(detatched from the gradient computation graph) 
using different masking ratios $r_1$ and $r_2$, respectively. 
The residual difference $\hat{x}_2-\hat{x}_1$ is used as supervision in the 
frequency domain for the detailed branch in loss 
$\mathcal{L}_{t}^{\mathrm{det}}$. 
An illustration can be found in \figref{fig:ex}.

A weighted sum of classification loss $\mathcal{L}_{t}^{\text{cls}}$, reconstruction loss $\mathcal{L}_{t}^{\text{rec}}$, and detail loss $\mathcal{L}_{t}^{\text{det}}$ is used as the overall loss for training:
\begin{equation}
  \begin{aligned}
    \label{eqn:all_loss}
    \mathcal{L}_{t} &=  \lambda_{\mathrm{cls}} \mathcal{L}_{t}^{\mathrm{cls}} + \lambda_{\mathrm{rec}} \mathcal{L}_{t}^{\mathrm{rec}} + \lambda_{\mathrm{det}} \mathcal{L}_{t}^{\mathrm{det}}.
  \end{aligned}
\end{equation}
Pseudocode for our method is given in Algorithm~\ref{alg:mae}.

\section{Experimental Results}

    \begin{figure*}[htp]
    \centering
    \includegraphics[width=.99\linewidth]{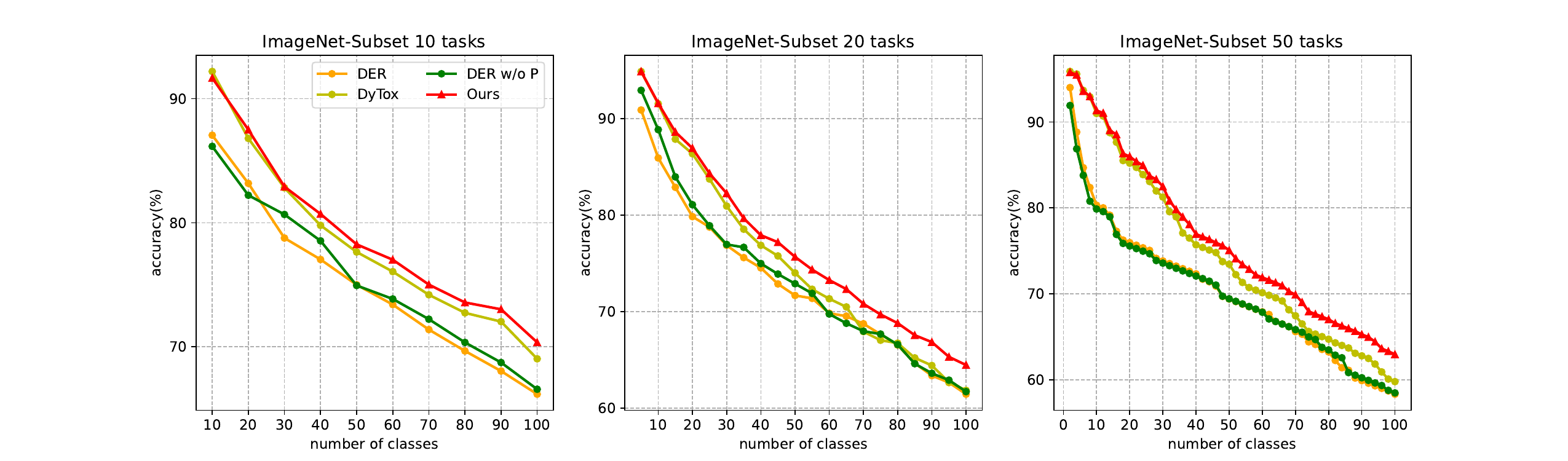}
    \vspace{-5pt}
	\caption{Performance evolution over incremental tasks on ImageNet-Subset. }
	\label{fig:subset}
\end{figure*}

  \begin{table*}[t]
	\centering
	\small
	\renewcommand{\arraystretch}{1.1}
  \setlength\tabcolsep{1.2mm}

	\begin{tabular}{ccccccccccccccc} 
\hline
	  \multirow{2}{*}{Method}& &\multicolumn{3}{c}{$N$=10}&&\multicolumn{3}{c}{$N$=20}&&\multicolumn{3}{c}{$N$=50} \\
	  \cline{3-5} \cline{7-9} \cline{11-13}
	   
	   &&Avg $\uparrow$&Last $\uparrow$&$F\downarrow$&&Avg $\uparrow$&Last $\uparrow$&$F\downarrow$&&Avg $\uparrow$&Last $\uparrow$&$F\downarrow$\\
	   \hline
BiC~\cite{wu2019large}&&64.96&55.07&31.32&&59.40&49.35&34.70&&53.75&44.56&40.23\\
PODNet~\cite{douillard2020podnet} &&63.44& 51.75&35.63&&55.11&45.37&41.70&&51.72&42.94&44.65\\
DER w/o P~\cite{yan2021dynamically}&&77.18& 66.70&14.86&&72.70&61.74&20.76&&70.44&58.87&24.20\\
DER~\cite{yan2021dynamically} &&76.12 &66.06&15.09&&72.56&61.51&20.46&&69.77&58.19&25.35\\
DyTox~\cite{douillard2022dytox}&&  77.15&69.10&14.66&& 73.13&61.87&17.32&& 71.51&60.02& 20.54\\ \hline
Ours&&  \textbf{79.54} &\textbf{70.29}& \textbf{12.04}&& \textbf{75.20}& \textbf{64.40}& \textbf{14.89}&& \textbf{74.42}& \textbf{62.87}&\textbf{17.22}\\
\hline
	\end{tabular} 
	\caption{Results on ImageNet-Subset in average and last phase accuracy (\%) and forgetting rate F (\%) on 10-, 20- and 50-task scenarios.}
	\label{tab:subset}
  \end{table*}
  
\begin{table}[t]
  \centering
  \small
  \renewcommand{\arraystretch}{1.1}
  \setlength\tabcolsep{1.2mm}
	\begin{tabular}{cccccccccc} \hline
	  \multirow{2}{*}{Method} & &\multicolumn{2}{c}{top-1}&\multicolumn{2}{c}{top-5} \\
	\cline{3-4} \cline{5-6}
		
		&&Avg $\uparrow$&Last $\uparrow$&Avg $\uparrow$&Last $\uparrow$\\
		\hline
iCaRL~\cite{rebuffi2017icarl}&&38.40&22.70&63.70&44.00\\
Simple-DER &&66.63&59.24& 85.62& 80.76\\
DER w/o P~\cite{yan2021dynamically}&&68.84&60.16& 88.17 &82.86\\
DER~\cite{yan2021dynamically} &&66.73&58.62&87.08 &81.89\\ 
DyTox~\cite{douillard2022dytox}&&71.29&63.34&88.59&84.49\\ \hline
Ours && \textbf{74.76}& \textbf{66.15}& \textbf{91.43}& \textbf{87.13} \\
\hline
\end{tabular} 
\caption{Results on ImageNet-Full for 10 incremental tasks. 
}
\label{tab:ifull}
\end{table}

\subsection{Benchmarks and Implementation}\label{ss:im}
\minisection{Datasets and settings.}
We experiment on three datasets: CIFAR-100~\cite{krizhevsky2009learning}, ImageNet-Subset, and ImageNet-Full~\cite{deng2009imagenet} to evaluate the performance of our approach. For CIFAR-100 and ImageNet-Subset, we test on 10-, 20- and 50-task scenarios, all with equal numbers of classes per task. We evaluate the 10-task setting for ImageNet-Full in which 100 new classes are included in each task. To measure the overall accuracy after all tasks during training, we report the average accuracy of learned tasks after each task and the accuracy of all tasks at the end of incremental learning.

\minisection{Implementation details.} We use the same network for all datasets. Models are trained from scratch to prevent data leakage with batch size of 1024 using Adam~\cite{kingma2015adam} with
initial learning rate 1e-4 and cosine decay. The loss weights from Eq.~\ref{eqn:all_loss} are set to $\lambda_{cls}=0.01$, $\lambda_{rec}=1.0$ and  $\lambda_{det}=1.0$.
The masking ratios are set to $r=0.75$, $r_1=0.75$, and $r_2=0.4$.
Each task is trained for 400 epochs. For exemplar-based methods from the literature, we store 20 samples for each classes (as is common practice).

We use 5 transformer blocks for the encoder and 1 for the decoder. All transformer blocks have an embedding dimension of 384 and 12 self-attention heads. This design differs from the original MAE as it is much more lightweight. We save image patches occupying the same amount of memory as other methods which store 20 full images per class. For example, we select 80 images and randomly save only 25\% patches from each using a masking ratio of 0.75 (thus only occupying the same space as 20 whole images). The detailed block is implemented with a 3-layer MLP keeping the dimension at 384. Further details about network architecture are given in the Supplementary Material.

\subsection{Comparison with the state-of-the-art}

In this section we compare our approach with the state-of-the-art, including DER~\cite{yan2021dynamically} and DyTox~\cite{douillard2022dytox}. In all plots and tables, ``DER w/o P'' denotes DER~\cite{yan2021dynamically} without pruning, therefore leading to more parameters being added across tasks. DyTox~\cite{douillard2022dytox} also uses a transformer architecture and we use the official codebase to reproduce its results.

\minisection{CIFAR-100.} We report results in average accuracy (Avg), the accuracy after the last task (Last), and average forgetting (F) in Table~\ref{tab:cf}. It is clear that under each setting our method outperforms others by a large margin. For longer task sequences, our bilateral MAE benefits from self-supervised reconstruction and richer replay data and it forgets much less compared to other methods. An overall accuracy curve is given in \figref{fig:cifar}. Using the same amount of replay storage, our method outperforms DyTox by about 6\% in accuracy after the last task in all three scenarios.

\minisection{ImageNet-Subset and ImageNet-Full.}
We report performance on ImageNet-Subset and ImageNet-Full in Tables~\ref{tab:subset}~and~\ref{tab:ifull}, respectively. Our method outperforms DyTox~\cite{douillard2022dytox} by 1.19\%, 2.53\%, and 2.85\% absolute gain in accuracy after the last task on the 10-, 20- and 50-task settings, respectively.  The higher average accuracy during each phase and lower forgetting also demonstrates the effectiveness of our method in alleviating forgetting. We also illustrate the performance evolution on ImageNet-Subset in \figref{fig:subset}. Our method has accuracy similar to DyTox in the first task, but in later tasks our method surpasses all others, especially for long task sequences. 
On the larger-scale ImageNet-Full, our Bilateral MAE significantly outperforms other methods by about 3\% in all metrics. 

\subsection{Ablation Study}

	


	


\begin{table}[tp]
	\centering
	\small
	\setlength\tabcolsep{1.3mm}
	\renewcommand{\arraystretch}{1.3}
	
	\begin{tabular}{cccc|c|c}
	\hline
	   Method&Replay&Reconstruction&Bilateral & Avg & Last  \\ \hline
	   Baseline&&&&73.40&62.31\\
	   Variants &\checkmark&&&75.88&64.35\\ 
&\checkmark&\checkmark&&77.48&66.54\\
&\checkmark&\checkmark&\checkmark&79.12&68.40\\

	   \hline
	\end{tabular}
	\caption{
	Ablative experiments on each component of our proposed method in the 10-task setting on CIFAR-100. 
	Replay denotes using generated data from MAE for replay, Reconstruction means applying the self-supervised reconstruction loss, and Bilateral indicates introduction the detailed branch of the MAE. 
	}
	\label{tab:fabl}
  \end{table}

\begin{table}[tp]
	\centering
	\small
	\setlength\tabcolsep{1.3mm}
	\renewcommand{\arraystretch}{1.3}
	
	\begin{tabular}{cc|c|c}
	\hline
	   $r$ &Data Source& Avg & Last  \\ \hline
	   0.60&Generated&77.50&67.37\\
	   0.75&Generated&79.12&68.40\\ 
	   0.90&Generated&77.12&67.02\\
N/A&Real&79.57&68.87\\
	   \hline
	\end{tabular}
	\caption{Ablation on the masking ratio and quality of generated data. Experiments are on CIFAR-100 in the 10-task setting and we report top-1 accuracy in \%. Data Source indicates whether replay is generated or real. In the last row we replay a selection of of real images equivalent in storage size to using $r_1=0.75$.
	}
	\label{tab:abr}
\end{table} 

\begin{table}[tp]
	\centering
	\small
	\setlength\tabcolsep{1.3mm}
	\renewcommand{\arraystretch}{1.3}
	
	\begin{tabular}{c|c|c}
	\hline
	   Domain& Avg & Last  \\ \hline
	   Spatial&77.45&65.93\\
Frequency&79.12&68.40\\

	   \hline
	\end{tabular}
	\caption{Ablative experiments on the detailed head. Experiments are on CIFAR-100 in the 10-task setting and we report the top-1 accuracy in \%. Domain denotes to which domain our loss is applied. 
	}
	\label{tab:abd}
  \end{table}

\begin{figure}[htp]
    \centering
    \includegraphics[width=.8\linewidth]{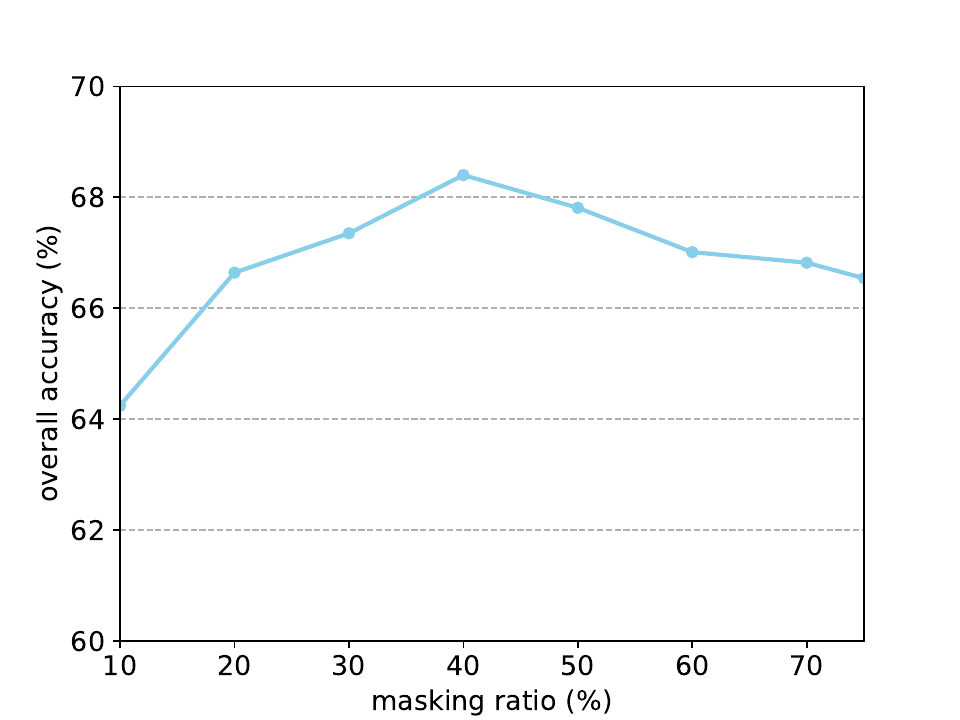}
	\caption{ Ablation on masking ratio $r_2$ in $\mathcal{L}_{t}^{\mathrm{det}}$. The other masking ratio $r_1$ is set to 75\% as the reference. }
	\label{fig:dms}
\end{figure}  

\minisection{Ablations on different components.} 
Our bilateral MAE consists of a self-supervised reconstruction task, generated data for replay, and a bilateral MAE branch for image- and embedding-level fusion. We ablate on these three factors in Table~\ref{tab:fabl}. 
These three major components in our method have different functions and they cooperate to boost performance compared with the baseline by about 6\%. We observe that: (a) More high-quality replay data has a direct contribution to the performance and masking ration of $r=0.75$ masking ratio yields 4$\times$ replay for the same storage cost as the baseline. (b) The reconstruction loss serves as effective self-supervision and improves performance by about 2\% in average accuracy. (c) The bilateral architecture works well by improving replay data generation quality and introducing image and embedding-level supervision. 


\newcommand{\addll}[1]{\includegraphics[width=0.22401\linewidth]{ours/#1.png}}
\newcommand{\addlo}[1]{\includegraphics[width=0.065\linewidth]{ours/#1.jpg}}
\begin{figure*}[t]
  \centering
  \small
  \setlength\tabcolsep{0.2mm}
    \renewcommand\arraystretch{0.6}
    \begin{tabular}{cccccc}
    \rotatebox{90}{\scriptsize{~~~~~~Pic 1}}&\addlo{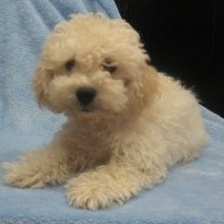}&\addll{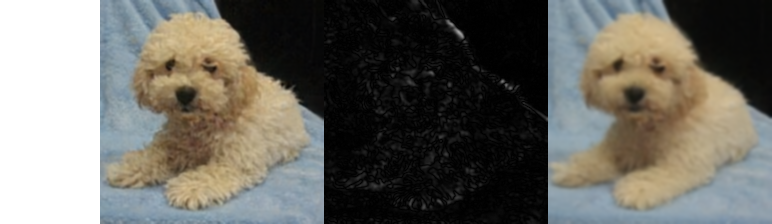}&\addll{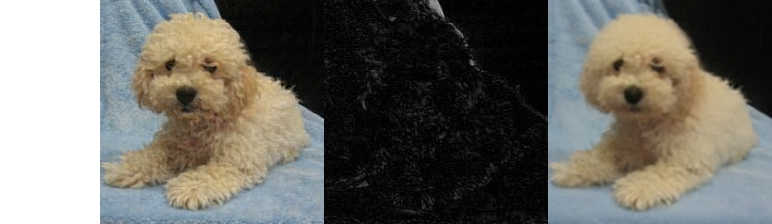}&\addll{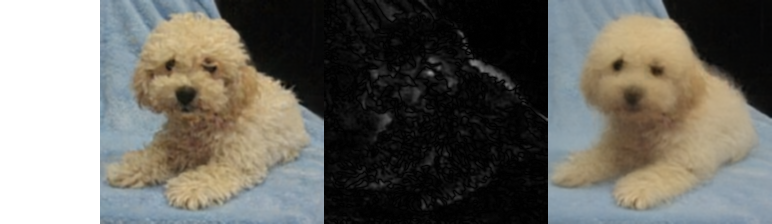}&\addll{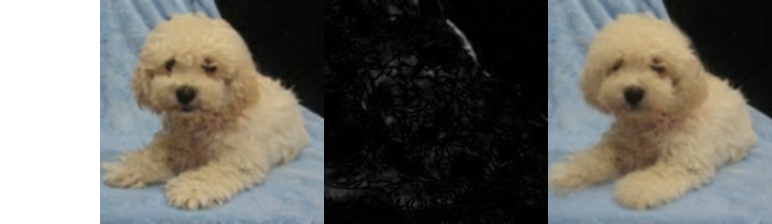}\\
    \rotatebox{90}{\scriptsize{~~~~~~~Pic 4}}&\addlo{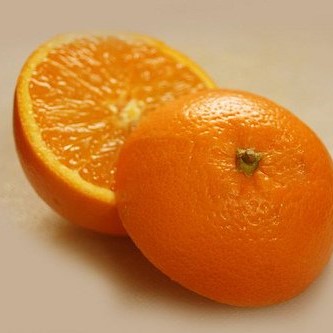}&\addll{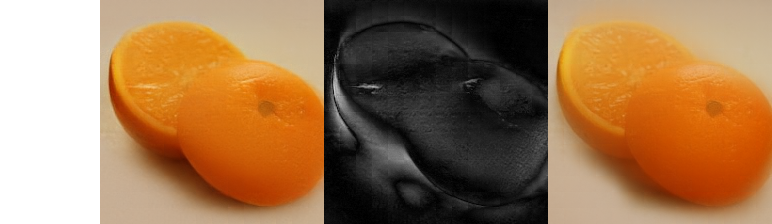}&\addll{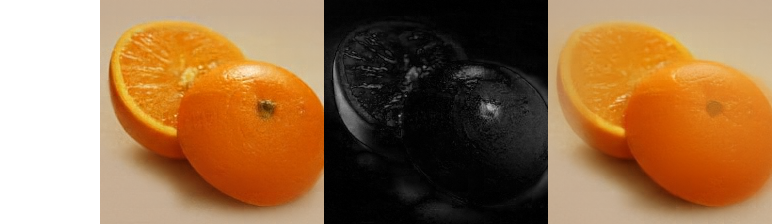}&\addll{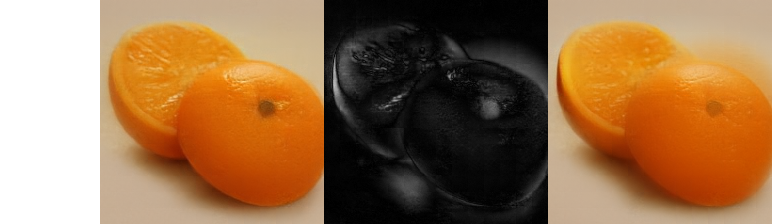}&\addll{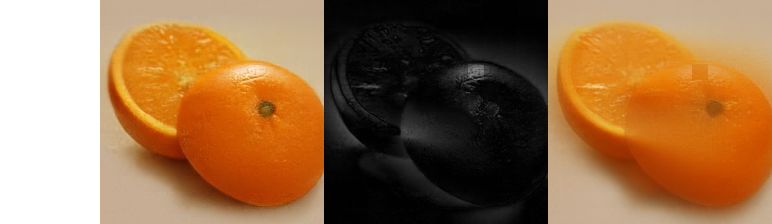}\\
    \rotatebox{90}{\scriptsize{~~~~~~~Pic 7}}&\addlo{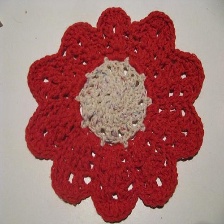}&\addll{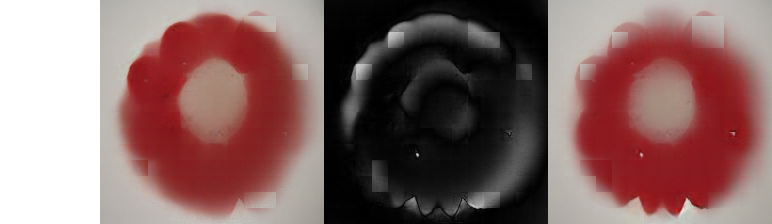}&\addll{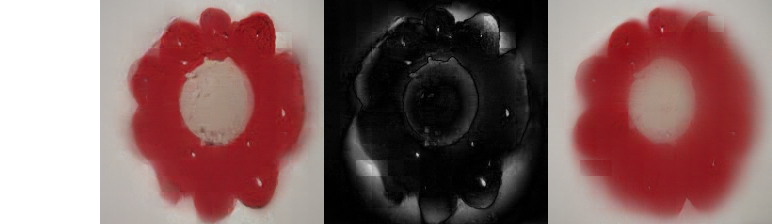}&\addll{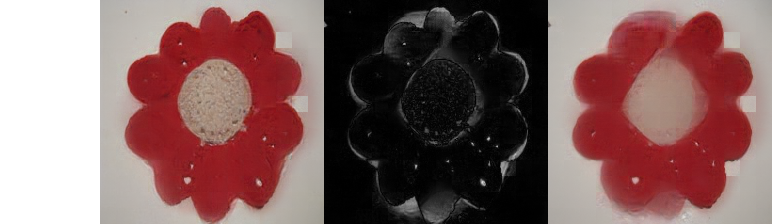}&\addll{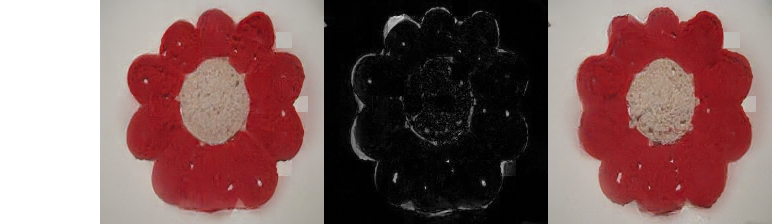}\\
    \rotatebox{90}{\scriptsize{~~~~Pic 10}}&\addlo{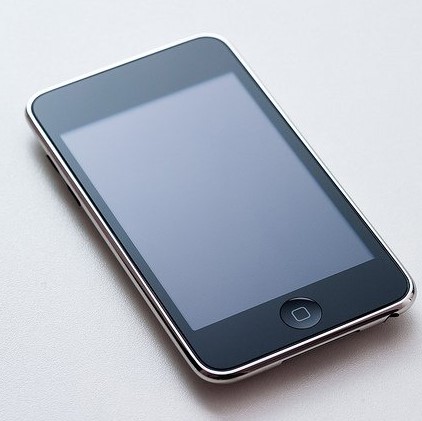}&\addll{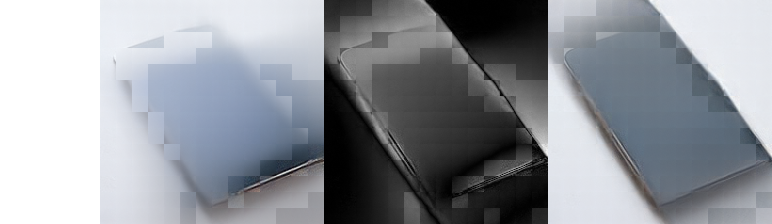}&\addll{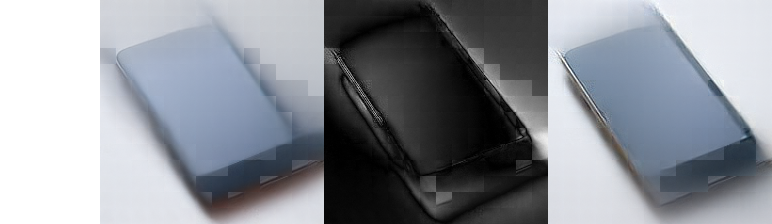}&\addll{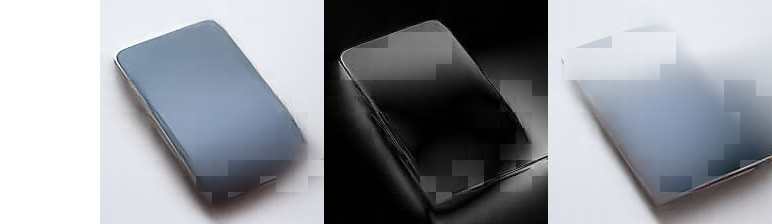}&\addll{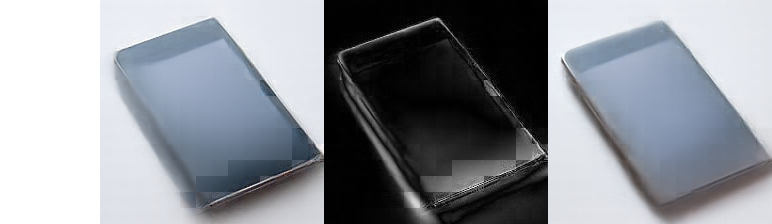}\\
    &Image&Task 1&Task 4&Task 7&Task 10
    \end{tabular}

		\caption{Reconstructions of images from ImageNet-Subset in the 10-task setting. Four original images were selected from tasks 1, 4, 6, and 10 and are shown on the left. In the remaining columns we show reconstructed images using our combined Bilateral MAE (left), only the detailed branch of our MAE (middle), and only the main branch of our MAE (right).
		}
		\label{fig:img}
\end{figure*}



\minisection{Masking ratio.} A key parameter of MAE~\cite{he2022masked} is the masking ratio $r$. There is trade-off on $r$: too large $r$ (e.g. 0.95) leads to poor reconstruction, which influences the quality of replay data and causes more serious forgetting. However, too small $r$ yields a limited extra amount of replay data (e.g. we can afford only about 11\% extra replay data when $r$ is 0.10). The results in Table~\ref{tab:abr} show that $r=0.75$ is a good trade-off for our bilateral MAE. To verify the quality of generated data for replay, also include results using original images for replay in place of generated ones. Results in rows 2 and 4 of Table~\ref{tab:abr} show that our method achieves good quality images with less than 0.5\% accuracy difference compared to replaying real images. 

\minisection{Frequency domain for the detailed loss.} We implemented the detailed loss by converting embeddings from the spatial domain to the frequency domain. This aims to allow concentration on high-frequency information which matches the learning objective of the detailed MAE branch. As shown in Table~\ref{tab:abd}, it is beneficial to make this conversion as it leads to a more than 2\% gain at the last task. 

\minisection{Ablation on $r_1$ and $r_2$ in the detailed loss.}
We set $r_1$ to 0.75 for all these experiments as a reference and we vary 
$r_2$ used for computing the ground truth of the detailed loss. The trade-off on $r_2$ is that large $r_2$ results in small differences between the reconstructed results from mask ratios $r_1$ and $r_2$ and therefore there is little information in the supervisory signal for the detailed loss and the impact of the detailed branch is reduced. 
On the other hand, too small $r_2$ (e.g. 0.10) maintains most of the residual part of reconstructed images from the main branch, which may bring weak supervision for the main branch and slow its training.
We show results for a range  of $r_2$ values in \figref{fig:dms}. These results show that an $r_2$ of about 0.40 is good for providing supervision to the detailed branch of our MAE. 

\minisection{Model and exemplar sizes.} To compare the effectiveness of different methods, normally models with the same or similar number of parameters are used with an equal number of exemplars. In our approach, we adapt the original MAE to be more lightweight and the number of parameters is comparable to or even smaller than DyTox, as shown in Table~\ref{tab:par} (last row). We set the masking ratio to 75\% by default and save 80 exemplars per class, and therefore the storage size for both models and exemplars is similar since our stored image patches require exactly the same as baselines using only 20 exemplars.

\begin{table}[tp]
	\centering
	\small
	\setlength\tabcolsep{1mm}
	\renewcommand{\arraystretch}{1.1}
	\begin{tabular}{ccccc}
	\hline
	  Method&Parameters (M)&Avg $\uparrow$&Last $\uparrow$&$F\downarrow$ \\ \hline
	  DER w/o P&112.27&75.36&65.22&15.02\\
	  DyTox&10.73&75.47&62.10&15.43\\
	  Ours (MLP size = 1536)&12.89&79.12& 68.40& 12.17\\
	  Ours (MLP size = 768)&9.35&78.36& 67.52& 12.90\\
	   \hline
	\end{tabular}
 \vspace{-0.1cm}
	\caption{
	Comparison of model sizes. We compare two versions of our Bilateral MAE and competing models. Experiments were conducted on CIFAR-100 with 10 tasks.
	}
 \vspace{-0.3cm}
	\label{tab:par}
\end{table} 

\minisection{Ablation on effective buffer size.}
In Table~\ref{tab:sample} we compare our method with Dytox using the same buffer size by masking input image patches in input to DyTox.
All three rows use the same memory size for storing exemplars. Using 80 exemplars with masking ratio of 75\%, DyTox (last row) achieves better performance than when using 20 full-image exemplars. It still performs worse than our approach, which shows that our performance gain does not simply come from the additional exemplars, but also from the integration of the MAE into our bilateral architecture with the Detailed Branch.


\minisection{Reconstruction analysis.}
We illustrate results of image reconstruction in the 10-task setting on ImageNet-Subset in \figref{fig:img}. 
Randomly selected images from tasks 1, 4, 7, and 10 are shown in the left column.  
Our bilateral MAE learns to reconstruct images in a task-agnostic way, which helps generate reasonable results for future tasks even before learning them. The detailed branch of our MAE learns to reconstruct high-frequency details complemetary to the main branch. 
The results from the main branch sometimes lack sample-specific characteristics, but with the help of our proposed detailed branch the reconstructions are more accurate and provide better generated replay data. 

\begin{table}[ht]
	\centering
	\small
	\renewcommand{\arraystretch}{1.1}
  \setlength\tabcolsep{1.1mm}

	\begin{tabular}{cccccc} 
\hline
	  Method&Buffer& Memory size & 25\% Patches & Images & Acc(\%)\\ \hline
   Ours&80&1x&\checkmark&&68.40\\

	  DyTox&20& 1x&&\checkmark&62.10\\
	  DyTox$\dagger$&80&1x &\checkmark&&65.46\\
\hline
	\end{tabular} 
	\caption{Ablation on effective buffer size with equal memory usage. Dytox$\dagger$ represents applying a mask ratio $r = 75\%$ directly to stored image exemplars in order to have the same number of exemplars and storage size for DyTox as in our setting. 
	}
 \vspace{-0.2cm}
	\label{tab:sample}
  \end{table}

\begin{figure}[t]
  \centering
  \includegraphics[width=\linewidth]{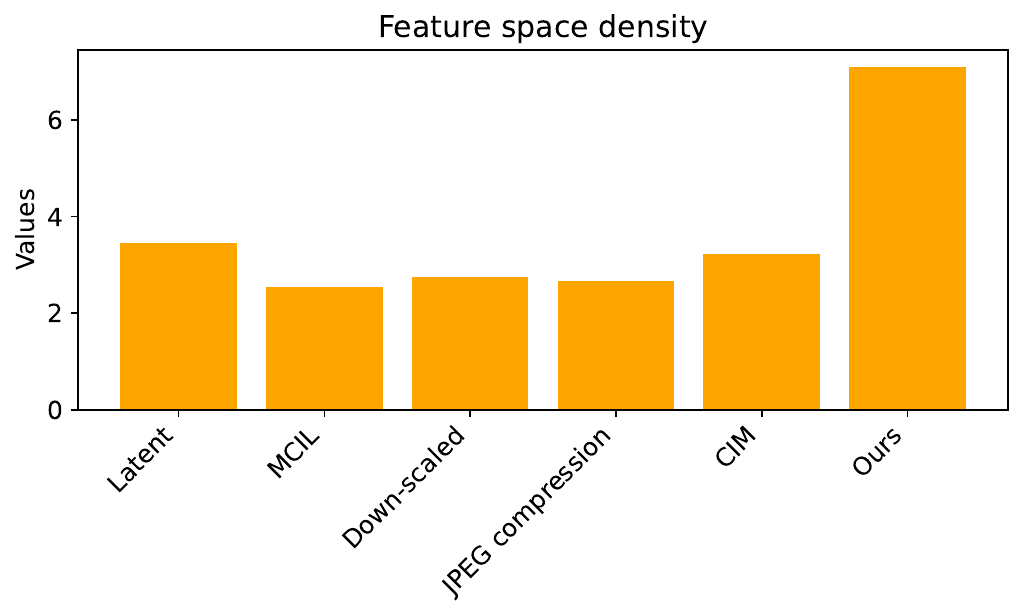}
  \vspace{-0.2cm}
  \caption{Comparison of feature space density $\pi$.}
  \label{fig:psd}
  \vspace{-0.6cm}
\end{figure}
\minisection{More generalizable representations help CIL.}
We calculated the feature space density metric~\cite{roth2020revisiting} for different methods following PASS~\cite{zhu2021prototype}: $\pi=\pi_{intra}/\pi_{inter}$, where $\pi_{intra}$ denotes the average cosine similarity within the same class and $\pi_{inter}$ denotes the one within different classes. Increased feature space density is associated with stronger generalization under data shift~\cite{zhu2021prototype}. We then compared feature space density after training all tasks, as shown in Figure~\ref{fig:psd} above. It is clear that Ours yields significantly higher density than other methods.

\begin{table}[htbp]
	\centering
	\small
	\renewcommand{\arraystretch}{1.1}
  \setlength\tabcolsep{1.2mm}
\resizebox{\linewidth}{!}{
\begin{tabular}{c|c|c|c}
\hline
\multicolumn{1}{c|}{\textbf{Metric}}& Memory&Avg$\uparrow$&Last$\uparrow$\\
\hline
Latent replay (CVPRW'20) \cite{liu2020generative} &- &62.44 &51.30 \\
MCIL (CVPR'20) \cite{liu2020mnemonics}&60& 63.25&53.12\\
Down-scaled (TNNLS'21) \cite{zhao2021memory}&60 &67.04 &55.40\\
JPEG compression (ICLR'22) \cite{wang2022memory}&60 &72.34 &61.32\\
CIM (CVPR'23) \cite{luo2023class} &60 &75.30 &63.05\\
Ours&60 & 79.12& 68.40\\
\hline
\end{tabular} 
}
\caption{Comparison of our bilateral MAE framework with other memory-efficient methods on the CIFAR-100 10-task setting. Memory indicates the required storage space per class (in KB). }
\label{tab:reconstruct}
\vspace{-0.2cm}
\end{table}
\minisection{Ablation with other efficient replay methods.} We compared the replay samples generated by MAE in our framework with a variety of memory-efficient methods based on latent replay~\cite{liu2020generative}, synthesized exemplars~\cite{liu2020mnemonics}, down-scaling~\cite{zhao2021memory}, JPEG image compression~\cite{wang2022memory}, and CIM~\cite{luo2023class} (foreground extraction and background compression). All these methods use the same amount of storage (except Latent Replay uses a GAN with 4.5M parameters), while our approach achieves consistently higher performance.
  
\section{Conclusions}
\label{sec:co}

In this work, we demonstrate that Masked Autoencoders are efficient incremental learners. Our approach stores random image patches as exemplars and it can reconstruct high-quality images from only partial information for replay. 
Furthermore, we propose a novel bilateral MAE architecture which further improves embedding diversity and reconstruction quality. Our Bilateral MAE approach significantly outperforms previous state-of-the-art methods for the same exemplar storage cost.

\minisection{Acknowledgements} This work is funded by  
NSFC (NO. 62225604, 62206135), 
and the Fundamental Research Funds for the Central Universities 
(Nankai Universitiy, 070-63233085). 
Computation is supported by the Supercomputing Center of Nankai University.

{\small
\bibliographystyle{ieee_fullname}
\bibliography{egbib}
}

\end{document}